\documentclass[runningheads]{llncs}
\usepackage{graphicx}
\usepackage{amsmath,amssymb} 
\usepackage{color}
\usepackage{multirow}
\usepackage{booktabs}
\usepackage[table]{xcolor}
\usepackage{tabularx,colortbl}
\usepackage{bbm}
\usepackage{bm}

\newcolumntype{b}{X}
\newcolumntype{s}{>{\hsize=.4\hsize\centering\arraybackslash}X}

\begin{document}
\title{VQA-E: Explaining, Elaborating, and Enhancing Your Answers for Visual Questions}

\titlerunning{VQA-E}
%
\author{Qing Li\inst{1} \and Qingyi Tao\inst{2,3} \and Shafiq Joty\inst{2} \and Jianfei Cai\inst{2} \and Jiebo Luo\inst{4}
}
%
\authorrunning{Qing Li, Qingyi Tao, Shafiq Joty, Jianfei Cai, and Jiebo Luo}
%

\author{Qing Li\textsuperscript{1}, Qingyi Tao\textsuperscript{2,3}, Shafiq Joty\textsuperscript{2}, Jianfei Cai\textsuperscript{2}, Jiebo Luo\textsuperscript{4}
}
\institute{\textsuperscript{1}University of Science and Technology of China,
	\textsuperscript{2}Nanyang Technological University,
	\textsuperscript{3}NVIDIA AI Technology Center,
	\textsuperscript{4}University of Rochester
}
\maketitle
\begin{abstract}
	Most existing works in visual question answering (VQA) are dedicated to improving the accuracy of predicted answers, while disregarding the explanations. We argue that the explanation for an answer is of the same or even more importance compared with the answer itself, since it makes the question answering process more understandable and traceable. 
	To this end, we propose a new task of VQA-E (VQA with Explanation), where the models are required to generate an explanation with the predicted answer. 
	We first construct a new dataset, and then frame the VQA-E problem in a multi-task learning architecture. Our VQA-E dataset is automatically derived from the VQA v2 dataset by intelligently exploiting the available captions.
	We also conduct a user study to validate the quality of the synthesized explanations . We quantitatively show that the additional supervision from explanations can not only produce insightful textual sentences to justify the answers, but also improve the performance of answer prediction. Our model outperforms the state-of-the-art methods by a clear margin on the VQA v2 dataset. 
\end{abstract}

\section{Introduction}
\begin{figure*}[]
	\begin{center}
		\includegraphics[width=0.6\linewidth]{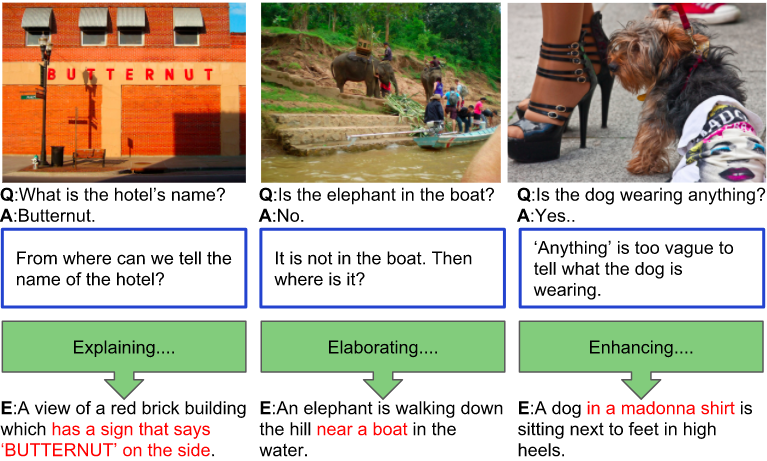}
	\end{center}
	\caption{VQA-E provides insightful information that can explain, elaborate or enhance  predicted answers compared with the traditional VQA task. Q=Question, A=Answer, E=Explanation. (Left) From the answer, there is no way to trace the corresponding visual content to tell the name of the hotel. The explanation clearly points out where to look for the answer. (Middle) The explanation provides a real answer to the aspect asked. (Right) The word ``anything'' in the question refers to a vague concept without specific indication. The answer is enhanced by the ``madonna shirt'' in the explanation. }\label{fig:eee_def}
\end{figure*}

In recent years, visual question answering (VQA) has been widely studied by researchers in both computer vision and natural language processing communities~\cite{antol2015vqa,zhu2016visual7w,goyal2017making,wu2016value,yang2016stacked,gurari2018vizwiz}. Most existing works perform VQA by utilizing attention mechanism and combining features from two modalities for predicting answers. 

Although promising performance has been reported, there is still a huge gap for humans to truly understand the model decisions without any explanation for them. A popular way to explain the predicted answers is to visualize attention maps to indicate \textit{`where to look'}. The attended regions are pointed to trace the predicted answer back to the image content. However, the visual justification through attention visualization is implicit and it cannot entirely reveal what the model captures from the attended regions for answering the questions. There could be many cases where the model attends to right regions but predicts wrong answers. What's worse, the visual justification is not accessible to visually impaired people who are the potential users of the VQA techniques. Therefore, in this paper we intend to explore textual explanations to compensate for these weaknesses of visual attention in VQA.

Another crucial advantage of textual explanation is that it elaborates and enhances the predicted answer with more relevant information. As shown in Fig.~\ref{fig:eee_def}, a textual explanation can be a clue to justify the answer, or a complementary delineation that elaborates on the context of the question and answer, or a detailed specification about abstract concepts mentioned in the QA to enhance the short answer. Such textual explanations are important for effective communication since they provide feedbacks that enable the questioners to extend the conversation. Unfortunately, although textual explanations are desired for both model interpretation and effective communication in natural contexts, little progress has been made in this direction, partly because almost all the public datasets, such as VQA~\cite{antol2015vqa,goyal2017making}, COCO-QA ~\cite{ren2015image}, and Visual7W~\cite{zhu2016visual7w}, do not provide explanations for the annotated answers. 

In this work, we aim to address the above limitations of existing VQA systems by introducing a new task called VQA-E (VQA with Explanations). In VQA-E, the models are required to provide a textual explanation for the predicted answer. We conduct our research in two steps. First, to foster research in this area, we construct a new dataset with textual explanations for the answers. 
The VQA-E dataset is automatically derived from the popular VQA v2 dataset \cite{goyal2017making} by synthesizing an explanation for each image-question-answer triple. The VQA v2 dataset is one of the largest VQA datasets with over 650k question-answer pairs, and more importantly, each image in the dataset is coupled with five descriptions from MSCOCO captions~\cite{chen2015microsoft}. Although these captions were written without considering the questions, they do include some QA-related information and thus exploiting these captions could be a good initial point for obtaining  explanations free of cost. We further explore several simple but effective techniques to synthesize an explanation from the caption and the associated  question-answer pair. To relieve concern about the quality of the synthesized explanations, we conduct a comprehensive user study to evaluate a randomly-selected subset of the explanations. The user study results show that the explanation quality is good for most question-answer pairs while being a little inadequate for the questions asking for a subjective response or requiring common sense (pragmatic knowledge). Overall, we believe the newly created dataset is good enough to serve as a benchmark for the proposed VQA-E task.

To show the advantages of learning with textual explanations, we also propose a novel VQA-E model, which addresses both the answer prediction and the explanation generation in a multi-task learning architecture. Our dataset enables us to train and evaluate the VQA-E model, which goes beyond a short answer by producing a textual explanation to justify and elaborate on it.
Through extensive experiments, we find that the additional supervisions from explanations can help the model better localize the important image regions and lead to an improvement in the accuracy of answer prediction. Our VQA-E model outperforms the state-of-the-art methods in the VQA v2 dataset. 

\section{Related Work}
\subsubsection{Attention in Visual Question Answering.}
Attention mechanism is firstly used in machine translation \cite{bahdanau2014neural} and then is brought into the vision-to-language tasks \cite{xu2015show,you2016image,xu2016ask,yang2016stacked,lu2016hierarchical,ilievski2016focused,nam2017dual,yu2017multi,gu2017empirical,gu2018stack,yang2018shuffle}. The visual attention in the vision-to-language tasks is used to address the problem of ``where to look'' \cite{shih2016look}. In VQA, the question is used as a query to search for the relevant regions in the image. \cite{yang2016stacked} proposes a stacked attention model which queries the image for multiple times to infer the answer progressively. Beyond the visual attention, Lu \textit{et al.} \cite{lu2016hierarchical} exploit a hierarchical question-image co-attention strategy to attend to both related regions in the image and crucial words in the question. \cite{nam2017dual} proposes the dual attention network, which refines the visual and textual attention via multiple reasoning steps. Attention mechanism can find the question-related regions in the image, which can account for the answer to some extent. \cite{das2017human} has studied how well the visual attention is aligned with the human gaze. The results show that when answering a question, current attention-based models do not seem to be ``looking'' at the same regions of the image as humans do. Although attention is a good visual explanation for the answer, it is not accessible for visually impaired people and is somehow limited in real-world applications.

\subsubsection{Model with Explanations.} 
Recently, a number of works \cite{hendricks2016generating,park2018multimodal,li2018tell} have been done for explaining the decisions from deep learning models, which are typically black boxes due to the end-to-end training procedure. 
\cite{hendricks2016generating} proposes a novel explanation model for bird classification. However, their class relevance metrics are not applicable to VQA since there is no pre-defined semantic category for the questions and answers. Therefore, we build a reference dataset to directly train and evaluate models for VQA with explanations. The most similar work to ours is \textit{Multimodal Explanations} \cite{park2018multimodal} that proposes a multimodal explanation dataset for VQA, which is human-annotated and of high quality. In contrast, our dataset focuses on textual explanations and is built free of cost and over six times bigger (269,786 v.s. 41,817) than theirs.

\section{VQA-E Dataset}

We now introduce our VQA-E dataset. We begin by describing the process of synthesizing explanations from image descriptions for question-answer pairs, followed by dataset analysis and a user study to assess the quality of our dataset.

\subsection{Explanation Synthesis} \label{Sec:ES}
\begin{figure*}[t]
	\begin{center}
		\includegraphics[width=1\linewidth]{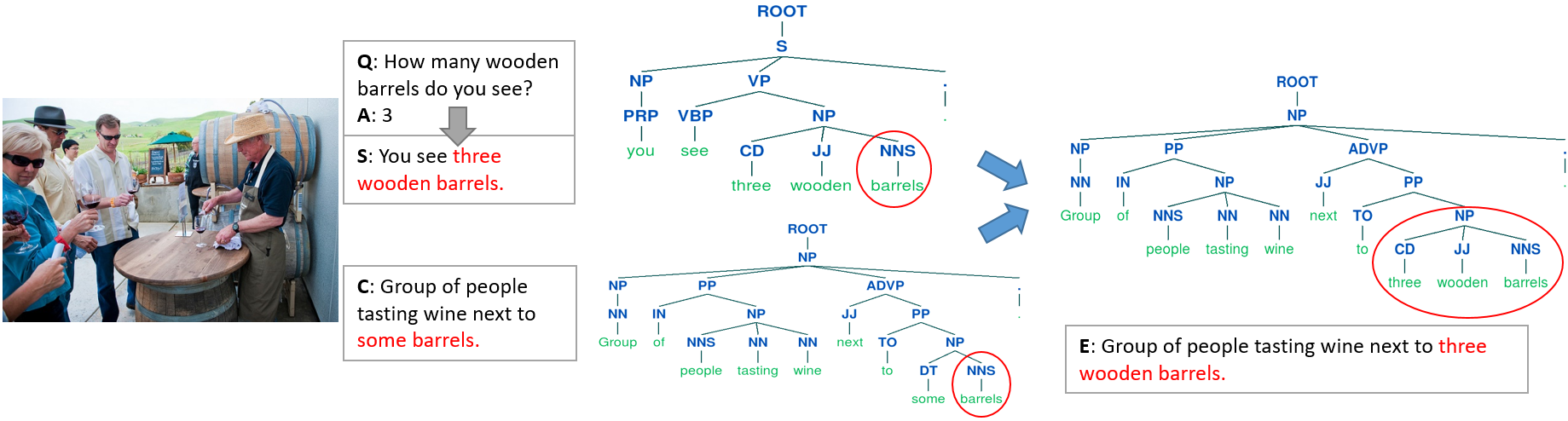}
	\end{center}
	\caption{An example of the pipeline to fuse the question (Q), the answer (A) and the relevant caption (C) into an explanation (E). Each question-answer pair is converted into a statement (S). The statement and the most relevant caption are both parsed into constituency trees. These two trees are then aligned by the common node. The subtree including the common node in the statement is merged into the caption tree to obtain the explanation.} \label{fig:sent_fusion}
\end{figure*}

\subsubsection{Approach.} The first step is to find the caption most relevant to the question and answer. Given an image caption $\mathcal{C}$, a question $\mathcal{Q}$ and an answer $\mathcal{A}$, we tokenize and encode them into GloVe word embeddings \cite{pennington2014glove}: $W_c = \{\bm{w}_1, ..., \bm{w}_{T_c}\}, W_q = \{\bm{w}_1, ..., \bm{w}_{T_q}\}, W_a = \{\bm{w}_1, ..., \bm{w}_{T_a}\}$, where $T_c, T_q, T_a$ are the number of words in the caption, question, and answer, respectively. We compute the similarity between the caption and question-answer pair as follows:
\begin{subequations}\label{eq:qac_simi}
	\begin{align}
	&s(\bm{w}_i, \bm{w}_j) = \frac{1}{2} (1+\frac{\bm{w}_i^T \bm{w}_j}{||\bm{w}_i||\cdot||\bm{w}_j||}) \\
	&S(\mathcal{Q},\mathcal{C}) = \frac{1}{T_q} \sum_{\bm{w}_i\in W_q}\max_{\bm{w}_j\in W_c} s(\bm{w}_i, \bm{w}_j) \\
	&S(\mathcal{A},\mathcal{C}) = \frac{1}{T_a} \sum_{\bm{w}_i\in W_a}\max_{\bm{w}_j\in W_c} s(\bm{w}_i, \bm{w}_j) \\
	&S(<\mathcal{Q}, \mathcal{A}>,\mathcal{C}) = \frac{1}{2}(S(\mathcal{Q},\mathcal{C}) + S(\mathcal{A},\mathcal{C}))
	\end{align}
\end{subequations}

For each question-answer pair, we find the most relevant caption, coupled with a similarity score. We have tried other more complex techniques like using Term Frequency and Inverse Document Frequency to adjust the weights of different words, but we find this simple mean-max formula in Eq.\eqref{eq:qac_simi} works better.

To generate a good explanation, we intend to fuse the information from both the question-answer pair and the most relevant caption. Firstly the question and answer are merged into a declarative statement. We achieve this by designing simple merging rules based on the question types and the answer types. Similar rule-based methods have been explored in NLP to generate questions from declarative statements \cite{Heilman:2010} (i.e.,  opposite direction). We then fuse this QA statement with the caption via aligning and merging their constituency parse trees. We further refine the combined sentence by a grammar check and correction tool to obtain the final explanation, and compute its similarity to the question-answer pair with Eq.~\ref{eq:qac_simi}. An example of our pipeline is shown in Fig.~\ref{fig:sent_fusion}.

\subsubsection{Similarity distribution.} 
\begin{figure*}[t]
	\begin{center}
		\includegraphics[width=1\linewidth]{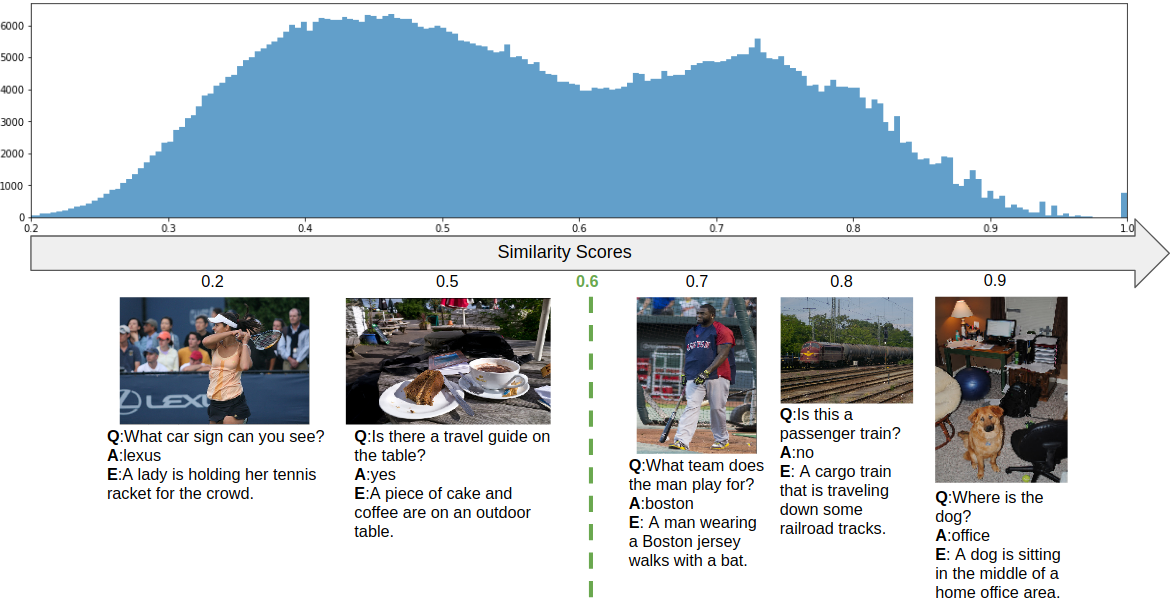}
	\end{center}
	\caption{Top: similarity score distribution. Bottom: illustration of VQA-E examples at different similarity levels.}\label{fig:word_embedding_similarity}
\end{figure*}

Due to the large size and diversity of questions, and the limited sources of captions for each image, it is not guaranteed that a good explanation could be generated for each Q\&A. The explanations with low similarity scores are removed from the dataset to reduce noise. We present some examples in Fig.~\ref{fig:word_embedding_similarity}. It shows a gradual improvement in explanation quality when the similarity scores increase. With some empirical investigation, we select a similarity threshold of 0.6 to filter out those noisy explanations. We also plot the similarity score histogram in Fig.~\ref{fig:word_embedding_similarity}. Interestingly, we observe a clear trough at 0.6 that makes the explanations well separated by this threshold. 

\subsection{Dataset Analysis}
\begin{figure*}[t]
	\begin{center}
		\includegraphics[width=1\linewidth]{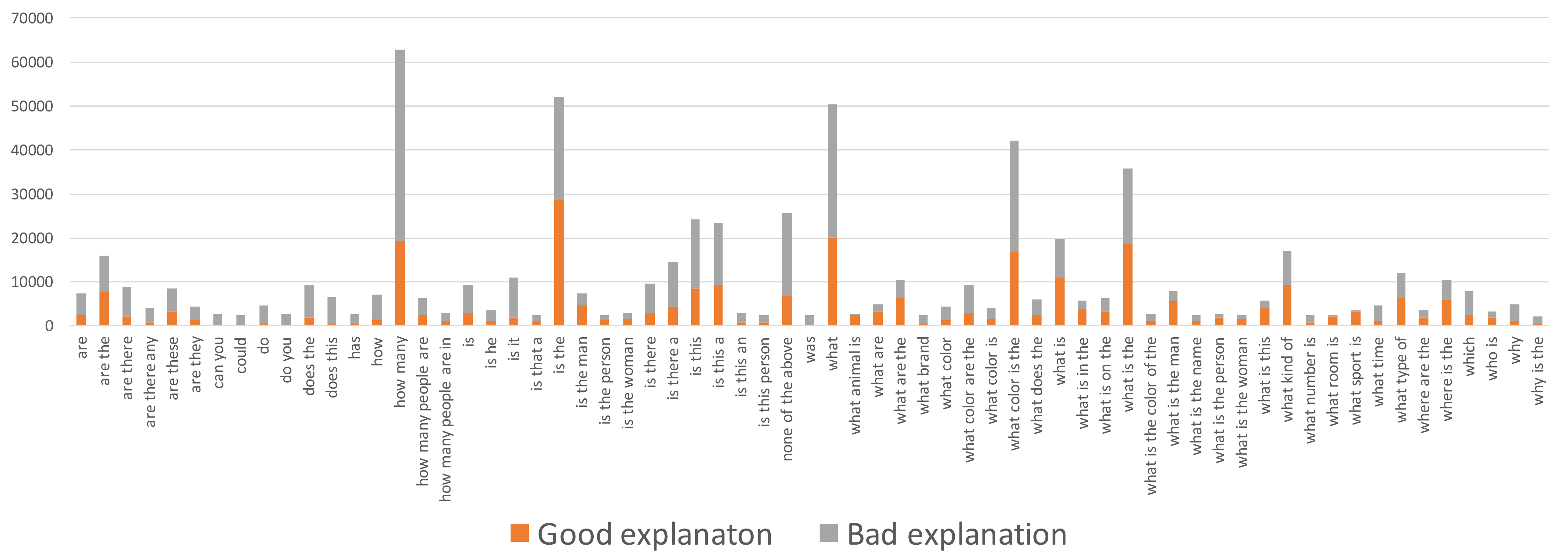}
	\end{center}
	\caption{Distribution of synthesized explanations by different question types.}\label{fig:word_relevance_distribution}
\end{figure*}

\begin{table}[t]
	\centering
	\small
	\caption{Statistics for our VQA-E dataset.}
	\label{tb:dataset_stats}
	\begin{tabularx}{1\textwidth}{l|c|@{\extracolsep{\fill}}cccccc}
		
		\toprule
		Dataset & Split & \#Images & \#Q\&A  & \#E & \#Unique Q & \#Unique A & \#Unique E\\
		\hline
		\multirow{3}{*}{\textbf{VQA-E}}   & Train & 72,680    & 181,298 & 181,298        & 77,418      & 9,491       & 115,560               \\
		& Val   & 35,645    & 88,488  & 88,488         & 42,055      & 6,247       & 56,916                \\
		& Total & 108,325   & 269,786 & 269,786        & 108,872     & 12,450      & 171,659               \\
		\hline
		\multirow{3}{*}{\textbf{VQA-v2}}  & Train & 82,783    & 443,757 & 0             & 151,693     & 22,531      & 0                    \\
		& Val   & 40,504    & 214,354 & 0             & 81,436      & 14,008      & 0                    \\
		& Total & 123,287   & 658,111 & 0             & 215,076     & 29,332      & 0                   \\
		\bottomrule
	\end{tabularx}
\end{table}

In this section, we analyze our VQA-E dataset, particularly the automatically synthesized explanations. Out of 658,111 existing question-answer pairs in original VQA v2 dataset, our approach generates relevant explanations with high similarity scores for 269,786 QA pairs (41\%). More statistics about the dataset are given in Table \ref{tb:dataset_stats}.

We plot the distribution of the number of synthesized explanations for each question type in Fig.~\ref{fig:word_relevance_distribution}. While looking into different question types, the percentage of relevant explanations varies from type to type.

\subsubsection{Abstract questions v.s. Specific questions.}
It is observed that the percentage of relevant explanations is generally higher for `is/are' and `what' questions than `how', `why' and `do' questions. This is because `is/are' and `what' questions tend to be related to specific visual contents which are more likely being described by image captions. In addition, a more specific question type could further help in the explanation generation. For example, for `what sport is' and for `what room is' questions, our approach successfully generates explanations for 90\% and 87\% question and answer pairs, respectively. The rates of having good explanations for these types of questions are much higher than the general `what' questions (40\%).

\subsubsection{Subjective questions: Do you/Can you/Do/Could?}
\begin{figure*}[t]
	\begin{center}
		\includegraphics[width=0.8\linewidth]{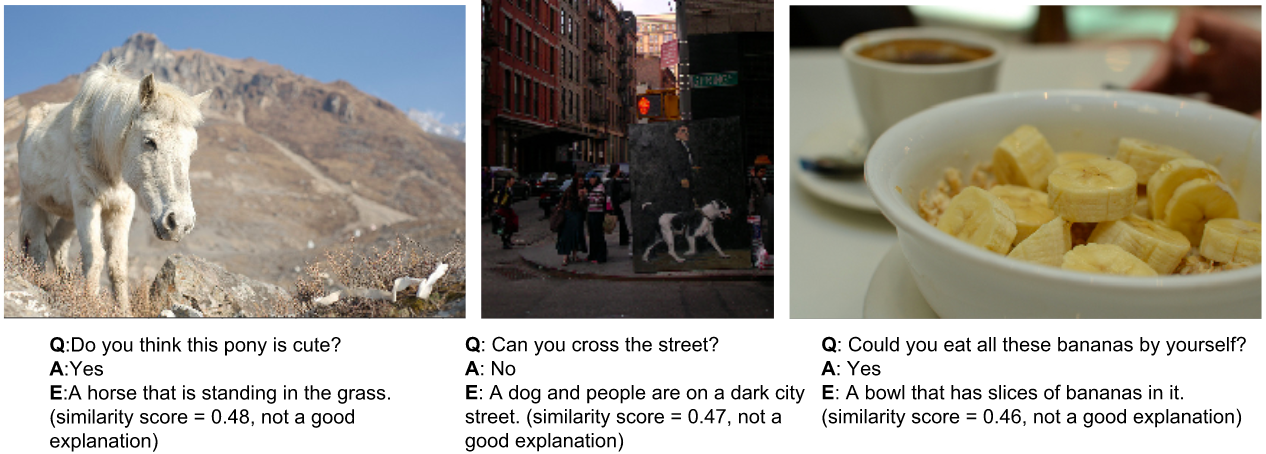}
	\end{center}
	\caption{Subjective examples: our method cannot handle the questions involving emotional feeling (left), commonsense knowledge (middle) or behavioral reasoning (right).}\label{fig:explanation_generation_bad_samples_do_can_vis}
\end{figure*}

The existing VQA datasets involve some questions that require subjective feeling, logical thinking or behavioral reasoning. These questions often fall in the question types starting with `do you', `can you', `do', `could', and etc. For these questions, there may be underlying clues  from the image contents but the evidence is usually opaque and indirect and thus it is hard to synthesize a good explanation. We illustrate examples of such questions in Fig. \ref{fig:explanation_generation_bad_samples_do_can_vis} and the generated explanations are generally inadequate to provide relevant details regarding the questions and answers. 

Due to the inadequacy in handling the above mentioned cases, we only achieve small percentages of good explanations for these question types. The percentages of `do you', `can you', `do' and `could' questions are 4\%, 5\%, 13\% and 6\% respectively which are far below the average 41\%. 

\subsection{Dataset Assessment -- User Study}
It is not easy to use quantitative metrics to evaluate whether the synthesized explanations can provide valid, relevant and complementary information to the answers of the visual questions. Therefore, we conduct a user study to assess our VQA-E dataset from human perspective. Particularly, we measure the explanation quality from four aspects: \textit{fluent}, \textit{correct}, \textit{relevant}, \textit{complementary}. 

\textit{Fluent} measures the fluency of the explanation. A fluent explanation should be correct in grammar and idiomatic in wording. The \textit{correct} metric indicates whether the explanation is correct according to the image content. The \textit{relevant} metric assesses the relevance of an explanation to the question and answer pair. If an explanation is relevant, users should be able to infer the answer from the explanation. This metric is important to measure whether the proposed word embedding similarity can effectively select and filter the explanations. Through the user study, we evaluate the relevance of explanations from human understanding to verify whether the synthesized explanations are closely tied to their corresponding QA pairs. Last but not least, we evaluate whether an explanation is \textit{complementary} to the answer. It is essential that the explanation can provide complementary details to the abbreviate answers so that visual accordance between the answer and the image could be enhanced.


\subsubsection{Evaluation results summary.}\label{sec:eval_result}
We show the human evaluation results in Table. \ref{tb:user_study_results}. Since the synthesized explanations are derived from existing human annotated captions, their average fluency and correctness scores are both close to 5. More importantly, their relevance and complementariness scores are both above 4, which indicates that the overall quality of the explanations is good from human perspective. These two metrics differentiate a general caption of an image and our specific explanation dedicated for a visual question-answer pair.


\begin{table}[t]
	\centering
	\caption{User assessment results for the synthesized explanation, the most similar caption, the random caption, and the generated explanation. To avoid bias, they are evaluated jointly and in each sample, their order is shuffled and unknown to users. They are assessed by the human evaluators in 1-5 grades: 1-very poor, 2-poor, 3-barely acceptable, 4-good, 5-very good. Here we show the average scores of 2,000 questions.}
	\label{tb:user_study_results}
	\begin{tabular}{l|cccc}
		\toprule
		& Fluent & Correct & Relevant & Complementary \\
		\hline
		Synthesized Explanation & 4.89   & 4.78    & \textbf{4.23}     & \textbf{4.14}  \\
		\hline
		Most Similar Caption & \textbf{4.97}   & 4.91    & 2.72      & 2.87           \\
		\hline
		Random Caption & 4.93   & \textbf{4.92}   &  1.91   &   2.12        \\
		\hline
		Generated Explanation (QI-AE) & 3.89  & 3.67  & 3.24 & 3.11           \\
		\bottomrule
	\end{tabular}
\end{table}





\section{Multi-task VQA-E Model}
\begin{figure*}[h]
	\centering {\includegraphics[width=0.9\textwidth]{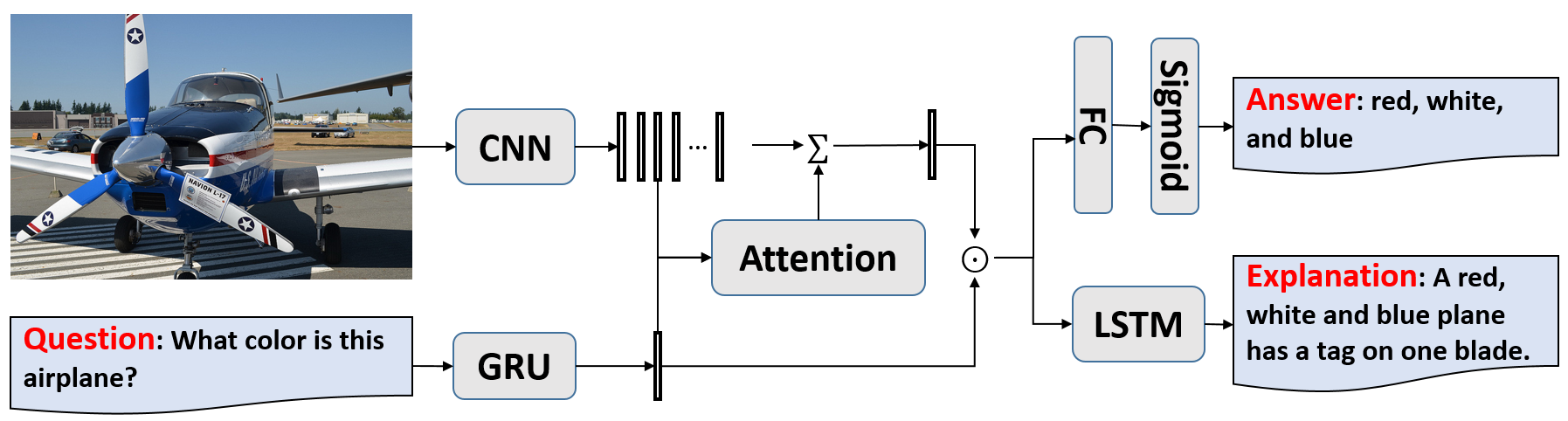}} 
	\caption{An overview of the multi-task VQA-E network. Firstly, an image is represented by a pre-trained CNN, while the question is encoded via a single-layer GRU. Then the image features and question features are input to the Attention module to obtain image features for question-guided regions. Finally, the question features and attended image features are used to simultaneously predict an answer and generate an explanation.}\label{fig:Multi-task}
\end{figure*}

Based on the well-constructed VQA-E dataset, in this section, we introduce the proposed multi-task VQA-E model. Fig.~\ref{fig:Multi-task} gives an overview of our model. Given an image $\mathcal{I}$ and a question $\mathcal{Q}$, our model can simultaneously predict an answer $\mathcal{A}$ and generate a textual explanation $\mathcal{E}$.

\subsection{Image Features} 
We adopt a pre-trained convolutional neural network (CNN) to extract a high-level representation $\phi$ of the input image $\mathcal{I}$:
\begin{equation}
\phi = \text{CNN}(\mathcal{I}) = \{\bm{v}_1, ..., \bm{v}_P\}
\end{equation}
where $\bm{v}_i$ is the feature vector of the $i^{th}$ image patch and $P$ is the total number of patches. We experiment with three types of image features:
\begin{itemize}
	\item \textbf{Global}. We extract the outputs of the final pooling layer (`pool5') of the ResNet-152~\cite{he2016deep} as global features of the image. For these image features, $P=1$, and visual attention is not applicable.
	\item \textbf{Grid}. We extract the outputs of the final convolutional layer (`res5c') of ResNet-152 as the feature map of the image, which corresponds to a uniform grid of equally-sized image patches. In this case, $P=7\times7=49$. 
	\item \textbf{Bottom-up}. \cite{anderson2017bottom} proposes a new type of image features based on object detection techniques. They utilize Faster R-CNN to propose salient regions, each with an associated feature vector from the ResNet-101. The bottom-up image features provide a more natural basis at the object level for attention to be considered. We choose $P=36$ in this case.
\end{itemize}

\subsection{Question Embedding}
The question $\mathcal{Q}$ is tokenized and encoded into word embeddings $W_q = \{\bm{w}_1, ..., \bm{w}_{T_q}\}$. Then the word embeddings are fed into a gated recurrent unit \cite{cho2014learning}:
$
\bm{q} = \text{GRU}(W_q).
$
We use the final state of the GRU as the representation of the question.

\subsection{Visual Attention}
We use the classical question-guided soft attention mechanism similar to most modern VQA models. For each patch in the image, the feature vector $\bm{v}_i$ and the question embedding $\bm{q}$ are firstly projected by non-linear layers to the same dimension. Next we use the Hadamard product (i.e., element-wise multiplication) to combine the projected representations and input to a linear layer to obtain a scalar attention weight associated with that image patch. The attention weights $\pmb{\tau}$ are normalized over all patches with softmax function. Finally, the image features from all patches are weighted by the normalized attention weights and summed into a single vector $\bm{v}$ as the representation of the attended image. The formulas are as follow and we omit the bias terms for simplicity:
\begin{equation}
\begin{split}
&{\tau}_i = \bm{w}^T~(\text{Relu}(W_v\bm{v}_i) \odot \text{Relu}(W_q\bm{q})) \\
&\pmb{\alpha} = \text{softmax}(\pmb{\tau}) \\
&\bm{v} = \sum_{i=1}^P \alpha_i \bm{v}_i
\end{split}
\end{equation}
Note that we adopt a simple one-glimpse, one-way attention, as opposed to complex schemes proposed by recent works~\cite{yang2016stacked,kazemi2017show,lu2016hierarchical}.

Next, the representations of the question $\bm{q}$ and the image $\bm{v}$ are projected to the same dimension by non-linear layers and then fused by a Hadamard product:
\begin{equation}
\bm{h} = \text{Relu}(W_{qh}\bm{q}) \odot \text{Relu}(W_{vh} \bm{v})
\end{equation}
where $\bm{h}$ is a joint representation of the question and the image, and then fed to the subsequent modules for answer prediction and explanation generation.

\subsection{Answer Prediction}
We formulate the answer prediction task as a multi-label regression problem, instead of a single-label classification problem in many other works. A set of candidate answers is pre-determined from all the correct answers in the training set that appear more than 8 times. This leads to $N=3129$ candidate answers. Each question in the dataset has $K=10$ human-annotated answers, which are sometimes not same, especially when the question is ambiguous or subjective and has multiple correct or synonymous answers. To fully exploit the disagreement between annotators, we adopt soft accuracies  as the regression targets. The accuracy for each answer is computed as:
\begin{equation} \label{eq:acc}
\begin{split}
\text{Accuracy}(a) &= \frac{1}{K} \sum_{k=1}^K \min(\frac{\sum_{1\leq j \leq K,j \neq k}\mathbbm{1}(a=a_j)}{3}, 1)\\
\end{split}
\end{equation}
Such soft target provides more information for training and is also in line with the evaluation metric.

The joint representation $\bm{h}$ is input into a non-linear layer and then through a linear mapping to predict a score for each answer candidate:
\begin{equation}
\hat{s} = \text{sigmoid}~(W_o~\text{Relu}~(W_f~\bm{h}))
\end{equation}
The sigmoid function squeezes the scores into $(0,1)$ as the probability of the answer candidate. Our loss function is similar to the binary cross-entropy loss while using soft targets:
\begin{equation}
L_{\text{vqa}} = -\sum_{i=1}^M\sum_{j=1}^N s_{ij}\log \hat{s}_{ij} + (1-s_{ij})\log(1-\hat{s}_{ij})
\end{equation}
where $M$ are the number of training samples and $\textbf{s}$ is the soft targets computed in Eq.\ref{eq:acc}. This final step can be seen as a regression layer that predicts the correctness of each answer candidate.

\subsection{Explanation Generation}
To generate an explanation, we adopt an LSTM-based language model that takes the joint representation $\bm{h}$ as input. Given the ground-truth explanation $\mathcal{E}=\{w_1, w_2, ..., w_{T_e}\}$, the loss function is:
\begin{equation}
\begin{split}
L_{\text{vqe}} &= -\log (p(\mathcal{E}|\bm{h}))\\
&= -\sum_{t=0}^{T_e} \log (p(w_t|\bm{h},w_1,...,w_{t-1}) )
\end{split}
\end{equation}

The final loss of multi-task learning is the sum of the VQA and VQE loss:
\begin{equation}
L = L_{\text{vqa}} + L_{\text{vqe}}
\end{equation}

\section{Experiments and Results}
\subsection{Experiment Setup}

\subsubsection{Model setting.}
We use 300 dimension word embeddings, initialized with pre-trained GloVe vectors~\cite{pennington2014glove}. For the question embedding, we use a single-layer GRU with 1024 hidden units. For explanation generation, we use a single-layer forward LSTM with 1024 hidden units. The question embedding and the explanation generation share the word embedding matrix to reduce the number of parameters. We use Adam solver with a fixed learning rate 0.01 and the batch size is 512. We use weight normalization \cite{salimans2016weight} to accelerate the training. Dropout and early stop (15 epochs) are used to reduce overfitting. 


\subsubsection{Model variants.}
We experiment with the following model variants:
\begin{itemize}
	\item \textbf{Q-E}: generating explanation from question only.
	\item \textbf{I-E}: generating explanation from image only.
	\item \textbf{QI-E}: generating explanation from question and image and only training the branch of explanation generation.
	\item \textbf{QI-A}: predicting answer from question and image and only training the branch of answer prediction.
	\item \textbf{QI-AE}: predicting answer and generating explanations, training both branches.
	\item \textbf{QI-AE(relevant)}: predicting answer and generating explanation and training both branches. The explanation used in this variant is the relevant caption obtained in the process of explanation synthesis in Section~\ref{Sec:ES}.
	\item \textbf{QI-AE(random)}: predicting answer and generating explanation and training both branches. The explanation is randomly selected from the ground-truth captions for the same image except the relevant caption.
	
\end{itemize}

\subsection{Evaluation of Explanation Generation}
\begin{table}[t]
	\centering
	\caption{Performance of explanation generation task on the validation split of the proposed VQA-E dataset, where B-N, M, R, and C are short for BLEU-N, METEOR, ROUGE-L, and CIDEr-D. All scores are reported in percentage (\%).}
	\begin{tabularx}{0.9\textwidth}{lc@{\extracolsep{\fill}}ccccccc}
		\toprule
		\textbf{Model}  & \textbf{Image Features} & \textbf{B-1}    & \textbf{B-2}    & \textbf{B-3}    & \textbf{B-4}    & \textbf{M}      & \textbf{C}      & \textbf{R} \\
		\midrule
		Q-E          & -               & 26.80           & 10.90           & 4.20            & 1.80            & 7.98            & 13.42           & 24.90 \\
		I-E          & Global          & 32.50           & 17.20           & 9.30            & 5.20            & 12.38           & 48.58           & 29.79 \\
		\midrule
		\multirow{3}[2]{*}{QI-E } & Global          & 34.70           & 19.30           & 11.00           & 6.50            & 14.07           & 61.55           & 31.87 \\
		& Grid            & 36.30           & 21.10           & 12.50           & 7.60            & 15.50           & 73.70           & 34.00 \\
		& Bottom-up       & 38.00           & 22.60           & 13.80           & 8.60            & 16.57           & 84.07           & 34.92 \\
		\midrule
		\multirow{3}[2]{*}{QI-AE} & Global          & 35.10           & 19.70           & 11.30           & 6.70            & 14.40           & 64.62           & 32.39 \\
		& Grid            & 38.30           & 22.90           & 14.00           & 8.80            & 16.85           & 87.04           & 35.16 \\
		& Bottom-up       & \textbf{39.30}  & \textbf{23.90}  & \textbf{14.80}  & \textbf{9.40}   & \textbf{17.37}  & \textbf{93.08}  & \textbf{36.33} \\
		\bottomrule
	\end{tabularx}%
	\label{tab:eg}
\end{table}%

In this section, we evaluate the task of explanation generation. Table.~\ref{tab:eg} shows the performance of all model variants on the validation split of the VQA-E dataset. First, the I-E model outperforms Q-E. This implies that it is easier to generate an explanation from only the image than from only the question, and this \textit{image bias} is contrary to the well-known \textit{language bias} in the VQA where it is easier to predict an answer from only the question than from only the image. Second, the QI-E models outperform both the I-E and Q-E by a large margin, which means that both the question and the image are critical for generating good explanations.  Attention mechanism is helpful for the performance and bottom-up image features are consistently better than grid image features. Finally, the QI-AE using bottom-up image features improves the performance further and achieves the best performance across all evaluation metrics. 
This shows that the supervision on the answer side is helpful for the explanation generation task, thus proving the effectiveness of our multi-task learning scheme.

\subsection{Evaluation of Answer Prediction}  
In this section, we evaluate the task of answer prediction, as shown in Table.~\ref{tab:ap}. Overall, the QI-AE models consistently outperform QI-A models across all question types. This indicates that forcing the model to explain can help it predict a more accurate answer. We argue that the supervision on explanation in QI-AE models can alleviate the headache of language bias in the QI-A models, because in order to generate a good explanation, the model has to fully exploit the image content, learn to attend to important regions, and explicitly interpret the attended regions in the context of questions. In contrast, during the training of QI-A models without explanations, when an answer can be guessed from the question itself, the model can easily get the loss down to zero by understanding the question only regardless of the image content. In this case, the training sample is not fully exploited to help the model learn how to attend to the important regions. Another observation from Table.~\ref{tab:ap} can further support our argument. The additional supervision on explanation produces a much bigger improvement on the attention-based models (Grid and Bottom-up) than the models without attention (Global). 

\begin{table}[t]
	\centering
	\caption{Performance of the answer prediction task on the validation split of VQA v2 dataset. Accuracies in percentage (\%) are reported.}\label{tab:ap}
	\begin{tabularx}{0.9\textwidth}{lc@{\extracolsep{\fill}}cccc}
		\toprule
		\textbf{Model}  & \textbf{Image features} & \textbf{All}    & \textbf{Yes/No} & \textbf{Number} & \textbf{Other} \\
		\midrule
		\multirow{3}[2]{*}{QI-A} & Global          & 57.26           & 77.19           & 39.73           & 46.74 \\
		& Grid            & 59.25           & 76.31           & 39.99           & 51.38 \\
		& Bottom-up       & 61.78           & 78.63           & 41.30           & 52.54 \\
		\midrule
		\multirow{3}[2]{*}{QI-AE} & Global          & 57.92           & 78.01           & 40.46           & 47.25 \\
		& Grid            & 60.57           & 78.35           & 39.36           & 52.66 \\
		& Bottom-up       & \textbf{63.51}  & \textbf{80.85}  & \textbf{43.02}  & \textbf{54.16} \\
		\midrule
		QI-AE(random) & Bottom-up       & 58.74           & 78.75           & 40.79           & 48.26 \\
		QI-AE(relevant) & Bottom-up       & 62.18           & 79.02           & 41.07           & 53.26 \\
		\bottomrule
	\end{tabularx}%
\end{table}%

\begin{table}[t]
	\centering
	\caption{Performance comparison with the state-of-the-art VQA methods on the test-standard split of VQA v2 dataset. \colorbox[rgb]{ .682,  .667,  .667}{BUTD-ensemble} is an ensemble of 30 models and it will not participate in ranking. Accuracies in percentage (\%) are reported.}\label{tab:ap-test}
	\begin{tabularx}{0.9\textwidth}{bssss}
		\toprule
		\textbf{Method} & \textbf{All}    & \textbf{Yes/No} & \textbf{Number} & \textbf{Other} \\
		\midrule
		Prior \cite{goyal2017making} & 25.98           & 61.20           & 0.36            & 1.17 \\
		Language-only \cite{goyal2017making} & 44.26           & 67.01           & 31.55           & 27.37 \\
		d-LSTM+n-I \cite{goyal2017making} & 54.22           & 73.46           & 35.18           & 41.83 \\
		MCB \cite{fukui2016multimodal,goyal2017making} & 62.27           & 78.82           & 38.28           & 53.36 \\
		BUTD \cite{teney2017tips,anderson2017bottom} & 65.67           & 82.20           & \textbf{43.90}  & 56.26 \\
		\rowcolor[rgb]{ .682,  .667,  .667} BUTD-ensemble \cite{teney2017tips,anderson2017bottom} & 70.34           & 86.60           & 48.64           & 61.15 \\
		\midrule
		Ours: QI-AE-Bottom-up     & \textbf{66.31}  & \textbf{83.22}  & 43.58           & \textbf{56.79} \\
		\bottomrule
	\end{tabularx}%
\end{table}%

QI-AE(random)-Bottom-up produces a much lower accuracy than QI-AE-Bottom-up, even lower than QI-A-Bottom-up. This implies that low-quality or irrelevant explanations might confuse the model, thus leading to a big drop in the performance. It also relieves the concern that the improvement is brought by learning to describe the image, rather than explaining the answer. This further substantiates the effectiveness of the additional supervision on explanation.

Table.~\ref{tab:ap-test} presents the performance of our method and the state-of-the-art approaches on the test-standard split of VQA v2 dataset. Our method outperforms the state-of-the-art methods over the answer types `Yes/No' and `Other' as well as in the overall accuracy, while producing a slightly lower accuracy over the answer type `Number' than BUTD \cite{teney2017tips,anderson2017bottom}.

\subsection{Qualitative Analysis}
\begin{figure*}[t] 
	\centering {\includegraphics[width=0.9\textwidth]{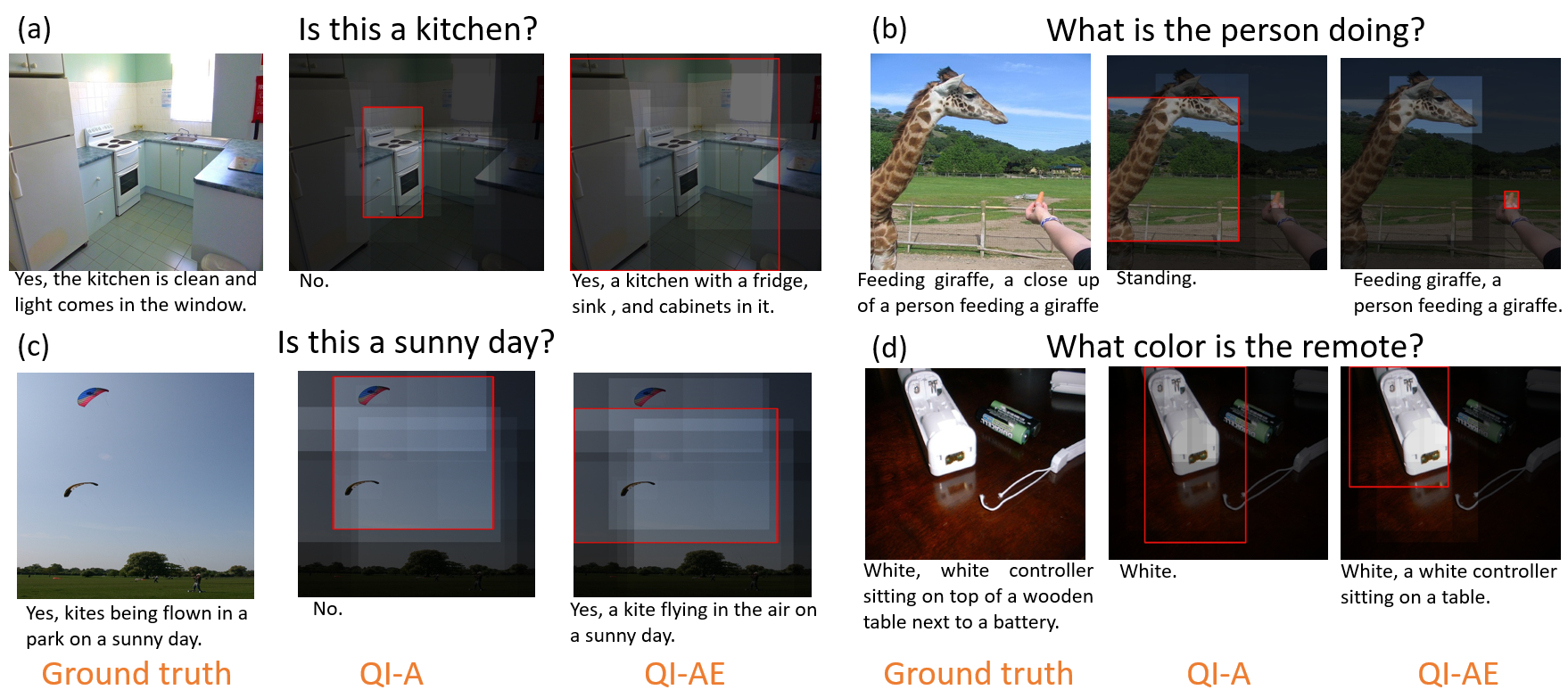}}
	\caption{Qualitative comparison between the QI-A and QI-AE models (both using bottom-up image features). We visualize the attention by rendering a red box over the region that has the biggest attention weight.}\label{fig:qualitative_examples}
\end{figure*}
In this section, we show qualitative examples to demonstrate the strength of our multi-task VQA-E model, as shown in Fig.\ref{fig:qualitative_examples}. Overall, the QI-AE model can generate relevant and complementary explanations for the predicted answers. For example, in the (a) of Fig.~\ref{fig:qualitative_examples}, the QI-AE model not only predicts the correct answer `Yes', but also provides more details in the `kitchen', i.e., `fridge', `sink', and `cabinets'. Besides, the QI-AE model can better localize the important regions than the QI-A model. As shown in the (b) of Fig.~\ref{fig:qualitative_examples}, the QI-AE model gives the biggest attention weight on the person's hand and thus predicts the right answer `Feeding giraffe', while the QI-A model focuses more on the giraffe, leading to a wrong answer `Standing'. In the (c), both QI-AE and  QI-E models attend to the right region, but these two models predict the opposite answers. This interesting contrast implies that the QI-AE model, which has to fully exploit the image content to generate an explanation, can better understand the attended region than the QI-A model that only needs to predict a short answer.

\section{Conclusions and Future Work}
In this work, we have constructed a new dataset and proposed a task of VQA-E to promote research on justifying answers for visual questions. Explanations in our dataset are of high quality for those visually-specific questions, while being inadequate for subjective ones whose evidences are indirect. For subjective questions, we will need extra knowledge bases to find good explanations for them.

We have also proposed a novel multi-task learning architecture for the VQA-E task. The additional supervision from explanations not only enables our model to generate reasons to justify predicted answers, but also brings a big improvement in the accuracy of answer prediction. Our VQA-E model is able to better localize and understand the important regions in images than the original VQA model. In the future, we will adopt more advanced approaches to train our model, like the reinforcement learning in image captioning~\cite{rennie2017self}.

\subsubsection*{Acknowledgements.}
We thank Qianyi Wu etc. for helpful feedback on the user study. This research is partially supported by NTU-CoE Grant and Data Science \& Artificial Intelligence Research Centre@NTU (DSAIR). Jiebo Luo would like to thank the support of Adobe and NSF Award \#1704309.

\bibliographystyle{splncs04}
\bibliography{egbib}

\begin{thebibliography}{10}
\providecommand{\url}[1]{\texttt{#1}}
\providecommand{\urlprefix}{URL }
\providecommand{\doi}[1]{https://doi.org/#1}

\bibitem{anderson2017bottom}
Anderson, P., He, X., Buehler, C., Teney, D., Johnson, M., Gould, S., Zhang,
  L.: Bottom-up and top-down attention for image captioning and visual question
  answering. CVPR  (2018)

\bibitem{antol2015vqa}
Antol, S., Agrawal, A., Lu, J., Mitchell, M., Batra, D., Lawrence~Zitnick, C.,
  Parikh, D.: Vqa: Visual question answering. In: ICCV (2015)

\bibitem{bahdanau2014neural}
Bahdanau, D., Cho, K., Bengio, Y.: Neural machine translation by jointly
  learning to align and translate. ICLR  (2014)

\bibitem{chen2015microsoft}
Chen, X., Fang, H., Lin, T.Y., Vedantam, R., Gupta, S., Doll{\'a}r, P.,
  Zitnick, C.L.: Microsoft coco captions: Data collection and evaluation
  server. CoRR  (2015)

\bibitem{cho2014learning}
Cho, K., Van~Merri{\"e}nboer, B., Gulcehre, C., Bahdanau, D., Bougares, F.,
  Schwenk, H., Bengio, Y.: Learning phrase representations using rnn
  encoder-decoder for statistical machine translation. arXiv preprint
  arXiv:1406.1078  (2014)

\bibitem{das2017human}
Das, A., Agrawal, H., Zitnick, L., Parikh, D., Batra, D.: Human attention in
  visual question answering: Do humans and deep networks look at the same
  regions? Computer Vision and Image Understanding  \textbf{163},  90--100
  (2017)

\bibitem{fukui2016multimodal}
Fukui, A., Park, D.H., Yang, D., Rohrbach, A., Darrell, T., Rohrbach, M.:
  Multimodal compact bilinear pooling for visual question answering and visual
  grounding. EMNLP  (2016)

\bibitem{goyal2017making}
Goyal, Y., Khot, T., Summers-Stay, D., Batra, D., Parikh, D.: Making the v in
  vqa matter: Elevating the role of image understanding in visual question
  answering. CVPR  (2017)

\bibitem{gu2018stack}
Gu, J., Cai, J., Wang, G., Chen, T.: Stack-captioning: Coarse-to-fine learning
  for image captioning. AAAI  (2018)

\bibitem{gu2017empirical}
Gu, J., Wang, G., Cai, J., Chen, T.: An empirical study of language cnn for
  image captioning. In: ICCV (2017)

\bibitem{gurari2018vizwiz}
Gurari, D., Li, Q., Stangl, A.J., Guo, A., Lin, C., Grauman, K., Luo, J.,
  Bigham, J.P.: Vizwiz grand challenge: Answering visual questions from blind
  people. CVPR  (2018)

\bibitem{he2016deep}
He, K., Zhang, X., Ren, S., Sun, J.: Deep residual learning for image
  recognition. In: CVPR (2016)

\bibitem{Heilman:2010}
Heilman, M., Smith, N.A.: Good question! statistical ranking for question
  generation. In: Human Language Technologies: The 2010 Annual Conference of
  the North American Chapter of the Association for Computational Linguistics.
  pp. 609--617. HLT '10, Association for Computational Linguistics,
  Stroudsburg, PA, USA (2010),
  \url{http://dl.acm.org/citation.cfm?id=1857999.1858085}

\bibitem{hendricks2016generating}
Hendricks, L.A., Akata, Z., Rohrbach, M., Donahue, J., Schiele, B., Darrell,
  T.: Generating visual explanations. In: ECCV. pp. 3--19. Springer (2016)

\bibitem{ilievski2016focused}
Ilievski, I., Yan, S., Feng, J.: A focused dynamic attention model for visual
  question answering. ECCV  (2016)

\bibitem{kazemi2017show}
Kazemi, V., Elqursh, A.: Show, ask, attend, and answer: A strong baseline for
  visual question answering. arXiv preprint arXiv:1704.03162  (2017)

\bibitem{li2018tell}
Li, Q., Fu, J., Yu, D., Mei, T., Luo, J.: Tell-and-answer: Towards explainable
  visual question answering using attributes and captions. arXiv preprint
  arXiv:1801.09041  (2018)

\bibitem{lu2016hierarchical}
Lu, J., Yang, J., Batra, D., Parikh, D.: Hierarchical question-image
  co-attention for visual question answering. In: NIPS. pp. 289--297 (2016)

\bibitem{nam2017dual}
Nam, H., Ha, J.W., Kim, J.: Dual attention networks for multimodal reasoning
  and matching. CVPR  (2017)

\bibitem{park2018multimodal}
Park, D.H., Hendricks, L.A., Akata, Z., Rohrbach, A., Schiele, B., Darrell, T.,
  Rohrbach, M.: Multimodal explanations: Justifying decisions and pointing to
  the evidence. In: CVPR (2018)

\bibitem{pennington2014glove}
Pennington, J., Socher, R., Manning, C.: Glove: Global vectors for word
  representation. In: EMNLP. pp. 1532--1543 (2014)

\bibitem{ren2015image}
Ren, M., Kiros, R., Zemel, R.: Image question answering: A visual semantic
  embedding model and a new dataset. NIPS  \textbf{1}(2), ~5 (2015)

\bibitem{rennie2017self}
Rennie, S.J., Marcheret, E., Mroueh, Y., Ross, J., Goel, V.: Self-critical
  sequence training for image captioning. CVPR  (2017)

\bibitem{salimans2016weight}
Salimans, T., Kingma, D.P.: Weight normalization: A simple reparameterization
  to accelerate training of deep neural networks. In: NIPS. pp. 901--909 (2016)

\bibitem{shih2016look}
Shih, K.J., Singh, S., Hoiem, D.: Where to look: Focus regions for visual
  question answering. In: ICCV. pp. 4613--4621 (2016)

\bibitem{teney2017tips}
Teney, D., Anderson, P., He, X., Hengel, A.v.d.: Tips and tricks for visual
  question answering: Learnings from the 2017 challenge. CVPR  (2018)

\bibitem{wu2016value}
Wu, Q., Shen, C., Liu, L., Dick, A., van~den Hengel, A.: What value do explicit
  high level concepts have in vision to language problems? In: CVPR (2016)

\bibitem{xu2016ask}
Xu, H., Saenko, K.: Ask, attend and answer: Exploring question-guided spatial
  attention for visual question answering. In: ECCV. pp. 451--466. Springer
  (2016)

\bibitem{xu2015show}
Xu, K., Ba, J., Kiros, R., Cho, K., Courville, A.C., Salakhutdinov, R., Zemel,
  R.S., Bengio, Y.: Show, attend and tell: Neural image caption generation with
  visual attention. In: ICML. vol.~14, pp. 77--81 (2015)

\bibitem{yang2018shuffle}
Yang, X., Zhang, H., Cai, J.: Shuffle-then-assemble: Learning object-agnostic
  visual relationship features. In: ECCV (2018)

\bibitem{yang2016stacked}
Yang, Z., He, X., Gao, J., Deng, L., Smola, A.: Stacked attention networks for
  image question answering. In: CVPR. pp. 21--29 (2016)

\bibitem{you2016image}
You, Q., Jin, H., Wang, Z., Fang, C., Luo, J.: Image captioning with semantic
  attention. In: CVPR (2016)

\bibitem{yu2017multi}
Yu, D., Fu, J., Mei, T., Rui, Y.: Multi-level attention networks for visual
  question answering. In: CVPR (2017)

\bibitem{zhu2016visual7w}
Zhu, Y., Groth, O., Bernstein, M., Fei-Fei, L.: Visual7w: Grounded question
  answering in images. In: CVPR. pp. 4995--5004 (2016)

\end{thebibliography}

%
%
%
%
\end{document}